\documentclass[letterpaper, 10 pt, conference]{ieeeconf}  

\IEEEoverridecommandlockouts                              

\usepackage[dvipsnames]{xcolor}

\usepackage{times}
\usepackage{amssymb}

\usepackage[utf8]{inputenc}

\usepackage{mathbbol}

\usepackage{esvect}

\usepackage[percent]{overpic}

\usepackage{tabu}

\usepackage{multirow}

\usepackage[overload]{empheq}

\usepackage{bm}

\definecolor{dg}{rgb}{0.1, 0.6, 0.2}       
\definecolor{b}{rgb}{0.0, 0.0, 1}          

\usepackage{epsfig}
\usepackage{float}
\usepackage{color}
\usepackage{booktabs} 
\usepackage{multirow} 
\usepackage{algpseudocode}
\usepackage[ruled,linesnumbered,vlined]{algorithm2e}

\usepackage{amsfonts}
\usepackage{amsmath}
\usepackage{mathrsfs}

\usepackage{amsthm}
\usepackage{mathtools}

\let\labelindent\relax
\usepackage{enumitem}       
\setenumerate[enumerate]{align=left}

\usepackage{graphicx}
\usepackage{subcaption}
\usepackage{caption}

\usepackage{balance}

\usepackage{multirow}

\usepackage[flushleft]{threeparttable}

\newcommand{\norm}[1]{\left\lVert#1\right\rVert}

\makeatletter
\newlength\tmp@\newlength\t@mp
\newcommand{\comp}[3]
  {\mathop{ \settowidth\tmp@{$\displaystyle\mathop{#1}^{#3}_{#2}$}
  \hbox to \tmp@{\hss \settowidth\t@mp{$\displaystyle #1$}\setlength\t@mp{.45\t@mp}
  $\displaystyle\mathop{#1}^{\hspace\t@mp #3}_{\hspace{-\t@mp}#2}$
  \hss} }}
\makeatother

\newcommand{\Int}[2]
{\int_{#1}^{#2}}

\DeclareMathOperator*{\argmin}{argmin}

\def\R{\mathbb{R}}

\def\pos{\mathbf{p}}

\def\rot{\mathbf{R}}
\def\tf{\mathbf{T}}
\def\trans{\mathbf{t}}
\def\SO{\mathrm{SO(3)}}

\def\SE{\mathrm{SE(3)}}

\def\Log{\mathrm{Log}}

\usepackage{xstring}
\newcommand{\shorturl}[1]
{
    \href{https://#1}{\nolinkurl{#1}}
}

\usepackage{array}
\usepackage[pagebackref=true,breaklinks=true,colorlinks=true,citecolor=blue,linkcolor=blue]{hyperref}
\usepackage{cleveref}
\usepackage{cite}

\hypersetup{
    colorlinks=true,
    linkcolor=blue,
    filecolor=blue,      
    urlcolor=blue,
    citecolor=blue,
}

\title{\LARGE \bf
HD-maps as Prior Information for Globally Consistent Mapping in GPS-denied Environments
}

\author{Waqas Ali$^{1}$, Patric Jensfelt$^{1}$ and Thien-Minh Nguyen$^{2}$ 
\thanks{$^{1}$Authors are with the Division of Robotics, Perception, and Learning (RPL), KTH Royal Institute of Technology, Stockholm 114 28, Sweden.
{\tt\small email: waqasali@kth.se}}%
\thanks{$^{2}$Thien-Minh Nguyen is with Centre for Advanced Robotics Technology Innovation (CARTIN), Nanyang Technological University, 50 Nanyang Ave, Singapore.}
}

\begin{document}

\maketitle
\thispagestyle{empty}
\pagestyle{empty}

\begin{abstract}

In recent years, prior maps have become a mainstream tool in autonomous navigation. However, commonly available prior maps are still tailored to control-and-decision tasks, and the use of these maps for localization remains largely unexplored.
To bridge this gap, we propose a lidar-based localization and mapping (LOAM) system that can exploit the common HD-maps in autonomous driving scenarios. Specifically, we propose a technique to extract information from the \textit{drivable area} and \textit{ground surface height} components of the HD-maps to construct 4DOF pose priors. These pose priors are then further integrated into the pose-graph optimization problem to create a globally consistent 3D map. Experiments show that our scheme can significantly improve the global consistency of the map compared to state-of-the-art lidar-only approaches, proven to be a useful technology to enhance the system's robustness, especially in GPS-denied environment. Moreover, our work also serves as a first step towards long-term navigation of robots in familiar environment, by updating a map. In autonomous driving this could enable updating the HD-maps without sourcing a new from a third party company, which is expensive and introduces delays from change in the world to updated map.

\end{abstract}

\section{INTRODUCTION}

For autonomous driving systems, the maps play a central role in most basic tasks such as localization and path planning. Their accuracy and global consistency is therefore crucial for successful operation.
Depending on the specific task and the sensors used, the format and use of the maps can be different.

For localization, the map is a product of the simultaneous localization and mapping (SLAM) process, where it is updated by marginalizing online sensor information such as camera or lidar data to serve as the reference for new estimation. In this paper, we focus on lidar-based systems and their corresponding maps. One of the critical issues of a SLAM algorithm is the accumulation of drift in the map and the estimate over time. Most modern techniques rely on loop closure detection to correct this drift. Another practical approach is to incorporate GPS data, which is widely adopted in industry. However, neither approach is perfect. For GPS-based methods, the signal can significantly degrade in urban areas due to non-line-of-sight or multi-path effects. On the other hand, loop closure detection may not be feasible if the path does not include any loop, and lidar-based loop closure detection may also easily fail due to the sparseness of lidar, 

\begin{figure}[thpb]
	\centering
	\includegraphics[scale=0.30]{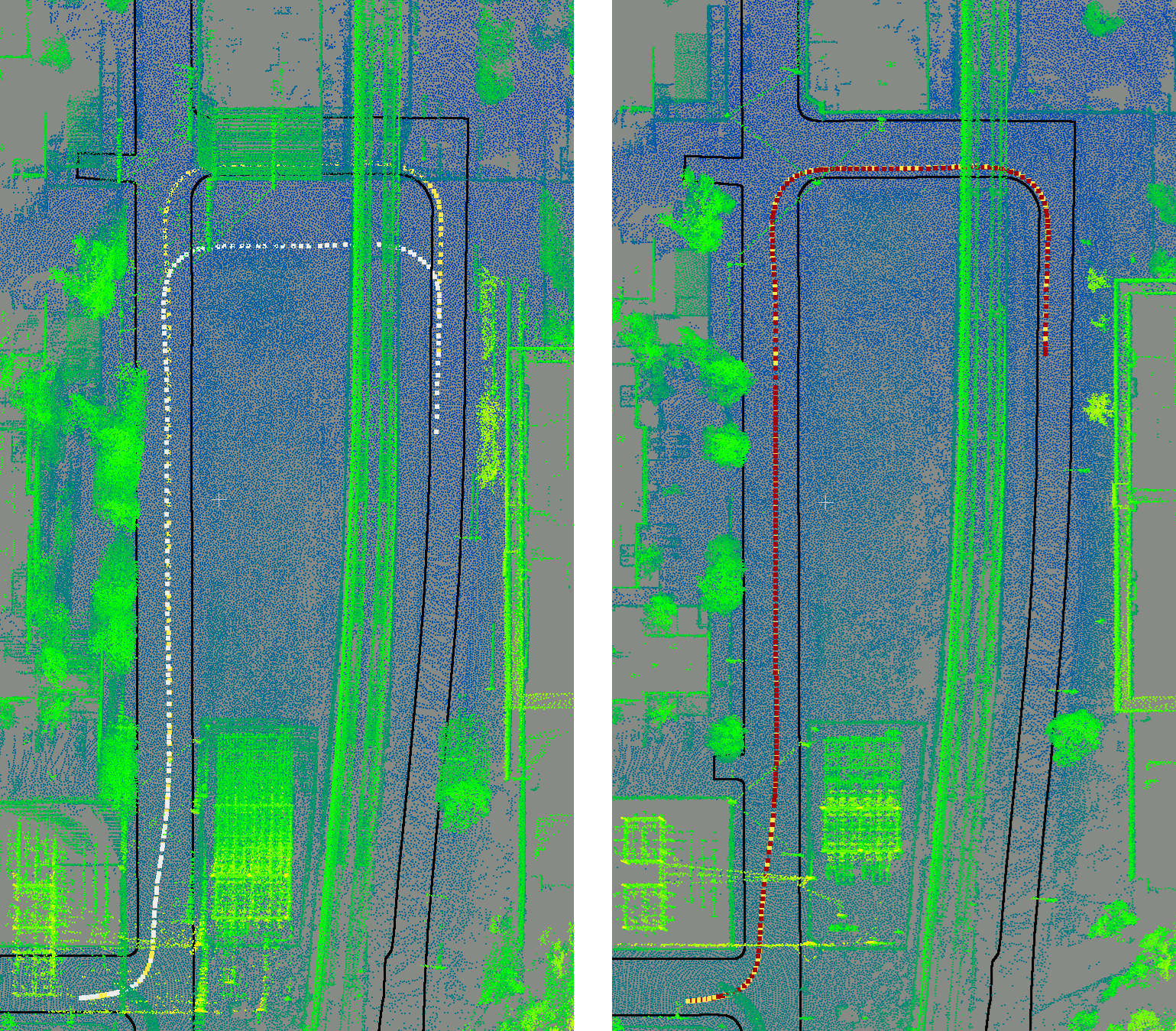}
	\caption{Left: Map built from ALOAM overlayed on the HD-map along with the ground truth (yellow) and ALOAM (white) trajectories where we see clear inconsistencies. Right: A more consistent map and the trajectory (red) generated from our system.}
	\label{aligned maps}
\end{figure}

\noindent
or that the scene has changed due to dynamic objects moving in and out. 

These issues will cause the resulting maps to distort, making then unusable for downstream tasks such as path planning, efficient routing or updating the HD map. In recent years some datasets such as MCD \cite{mcdviral2024}, Hilti \cite{zhang2022hilti} or Newer College \cite{ramezani2020newer} have provided dense survey-grade RGB dense prior maps from terrestrial survey scanners that can be used in the localization process. However these prior maps are tens to hundreds of GBs, which are not suitable for real time SLAM. Rather, the dense prior maps are used for registering lidar point cloud to generate dense ground truth.

Different from the internally built map of SLAM, in control and decision-making processes, such as in autonomous driving, we often rely on the map representing information such as drivable area, lane geometry and pedestrian crossings. Most modern autonomous vehicles leverage this information to complement the limitations of on-board sensors for path planning and localization support \cite{yu2014semiautomated, gandia2019autonomous}. A typical example of this kind is the HD-maps of Argoverse 2 Map Change Dataset, also known as the ``Trust, but Verify'' dataset (A2-TbV) \cite{wilson2023argoverse, lambert2022trust}, which provides information about drivable area, lanes and road surface height as vector maps. Such maps are only a few MBs in size, while containing all the key information required to support an autonomous navigation system. 

In this paper, we propose to utilize the HD-maps components as prior information to optimize the SLAM trajectory and generate a globally consistent map aligned with global coordinates. Our main contribution is to incorporate the drivable area vector map and ground surface height map to compute constraint used in pose-graph optimization. The proposed method is built over a traditional Lidar-based (LOAM framework) SLAM algorithm. Such a system ensure accurate localization and mapping result for long-term application while also making the system robust to challenging environments. This work is a first step towards enabling HD-map change detection and updates which is the task for which the dataset we use was developed.

The primary contributions of our works are as follows:

\begin{itemize}
	\item We propose a method to construct prior information for a SLAM algorithm from HD-map components, in our case the drive-able area vector map and ground height map.
	\item We introduce a pose-graph optimization scheme with HD-map-based factors to generate accurate a 6DOF trajectory and a globally consistent map.
	\item We demonstrate that our approach achieves significantly higher accuracy as compared to current state-of-the-art LOAM algorithms, proven by an extensive evaluation on Argoverse 2 Trust-but-Verify \cite{wilson2023argoverse} dataset. 	
\end{itemize}

\section{RELATED WORK}



Lidar-based SLAM algorithms are typically more robust and accurate than camera based versions which is evident from the many SLAM challenges \cite{helmberger2022hilti}. We will therefore focus on LIDAR based solutions. There has been tremendous advancement in such algorithms. Zhang and Singh \cite{zhang2017low} proposed LOAM, a feature-based Lidar odometry and mapping algorithm that can produce real-time accurate performance. To limit the computational requirements, they proposed a feature detection method that estimate the smoothness of the local surface and based on that can detect edge and planar features. Several modern Lidar based SLAM algorithms are built on the LOAM framework. Shan and Englot proposed Lego-LOAM \cite{legoLoam}, by optimizing the framework of LOAM. They used range image based segmentation to separate ground points in the scan and implement loop closure to improve the accuracy. Similarly, F-LOAM \cite{floam} presents a computationally efficient version of LOAM by applying a non-iterative two-step motion distortion compensation. Loam-livox \cite{loamlivox} is another extension of LOAM, which is proposed for Lidars with smaller FOV. Several Lidar based methods propose the integration of IMU data to get accurate performance in real-time. Shan et al.  \cite{shan2020lio} built a lidar SLAM algorithm using IMU pre-integration. Their method uses IMU data to provide pose prior for the scan to local map registration. Besides LOAM based algorithms, several direct methods are also present in literature. Xu et al. propose Fast-LIO \cite{xu2021fast, xu2022fast}, which utilizes the whole point cloud and the odometry is estimated by scan-to-map association, i.e., for each point to a local neighborhood. Chen et al. \cite{dlo} proposed a novel odometry algorithm that uses the whole point cloud with minimal pre-processing to produce accurate results. Nguyen et al. \cite{nguyen2023slict} proposed a surfel based SLAM algorithm. The method uses multi-scale surfels to build the global map by organizing the surfels in a tree-like structure. Such a global map can be updated incrementally and remove the computational requirements of recomputing at each update. 

While these SLAM methods have proven accurate performance in public datasets and real-world application, they can be made more robust to challenging conditions by utilizing some form of prior information. In SLAM long term drift is typically remove with loop closure constraints or with the addition of GPS data. Loop closure detection adds extra computational requirements and in cases where we don't return to previously visited scenes drift cannot be bounded. Another important aspect is the global consistency and aligning the map with world frame, which is critical for autonomous driving applications. For this purpose, the use of GPS data is common and several earlier work \cite{stephen2001development, sukkarieh1999high, krakiwsky1988kalman} on the outdoor localization and mapping focused on the used GPS information. But the main limitation of GPS data is the uncertainty and blockage due to environment conditions. Therefore, employing some form of prior map that can be associated to the Lidar point cloud map can improve the robustness and accuracy of the system. 

Lee et al. \cite{lee2007constrained} explored the use of road map network as priori information to improve the localization accuracy and fuse such map with the SLAM algorithms. 
Several methods have been presented which utilize Google maps or Openstreetmap \cite{haklay2008openstreetmap} to improve the system accuracy. Ruchti et al. \cite{Ruchti} extract a road network information from Openstreetmap and they propose a classification scheme to localize the 3D pose within such road network. Vysotska and Stachniss \cite{vysotska2016exploiting} proposed a Graph SLAM which utilized the Openstreetmap as prior information to generate constraints for pose graph optimization. Their approach extracts building information from the Openstreetmap and match it to the point cloud map built from lidar data. Suger and Burgard \cite{Suger} utilize the semantic road information to associate the Lidar trajectory with a global trajectory from Openstreetmaps. Most recently, Cho et al \cite{cho2022openstreetmap} proposed an Openstreetmap descriptor which is based on the distance to the building. The descriptor calculates distance both for lidar scan and for Openstreetmap, then comparing the distance can improve the localization accuracy. These methods rely on the buildings information from Openstreetmap and can only be applied when there is enough building structure present in the environment. On the other hand, our method only relies on the information of driveable area and doesn't need building structure, making it suitable for challenging environments. There is still limited work on using prior maps for generating globally consistent and accurate maps.  In our work, we propose to use the HD-map components including driveable area map and the ground surface height map. These components offer compact representation with minimal memory usage as these components only use a couple of Mega-bytes in size. An important aspect of employing vector maps is that our system doesn't rely on the structures present around the driveable area, thus the proposed approach can be applied to unstructured environments. Using such map representation will allow us to elevate the accuracy of traditional SLAM algorithms in challenging conditions and make such systems more suitable for real-world applications.

\section{Preliminaries}

\subsection{Notations}

For a physical quantity $\pos$, we use the hat notation $\hat{\pos}$ to denote the estimate of $\pos$, and the breve notation $\breve{\pos}$ to denote a measurement or prior.
We denote the vehicle's \textit{position} as $\trans \in \R^3$, and its \textit{orientation} $\rot \in \SO$. Hence, the vehicle's \textit{pose} $\tf \in \SE$ is defined by the tuple $\tf \triangleq (\rot, \trans)$. The residual between two $\SE$ poses is then defined as as follows:
\begin{equation}
    \tf_1 \boxminus \tf_2
    \triangleq
    \begin{bmatrix}
        \Log(\rot_2^{-1}\rot_1)\\
        \rot_2^{-1}(\trans_1 - \trans_2)
    \end{bmatrix}
    \in \R^6.
\end{equation}
In addition, we use the notation $\norm{r}^2_\Omega \triangleq r^\top\Omega r$, where $r\in \R^n$ and $\Omega\in\R^{n\times n}$

\begin{figure}
	\begin{subfigure}{\linewidth}
		\centering
		\includegraphics[width=\linewidth]{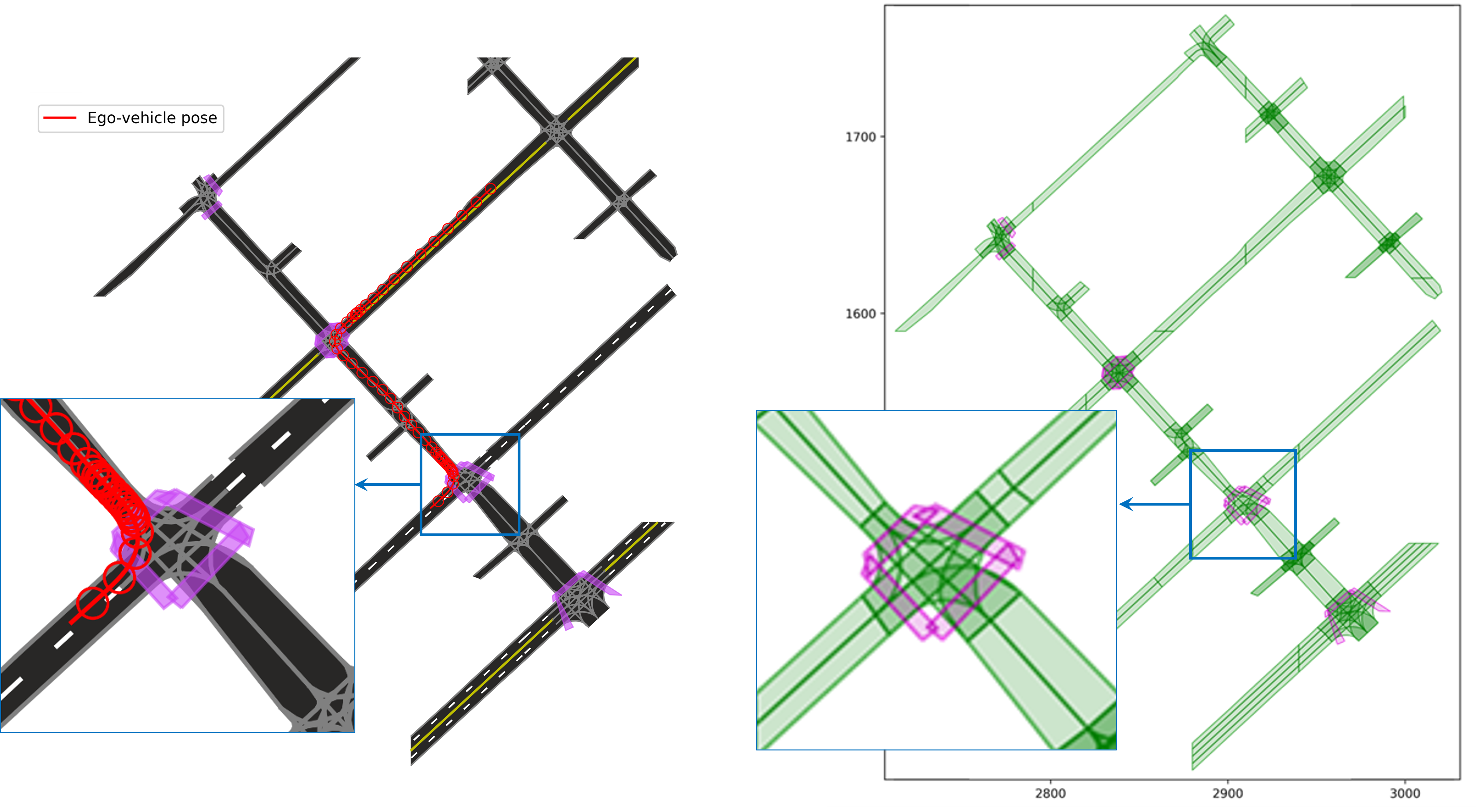}
            \caption{6-DOF pose on road map (left) and lane graph with pedestrian crossings (right)}
		\label{fig:hd-map-a}
	\end{subfigure}%
 
         \begin{subfigure}{\linewidth}
		\includegraphics[width=\linewidth]{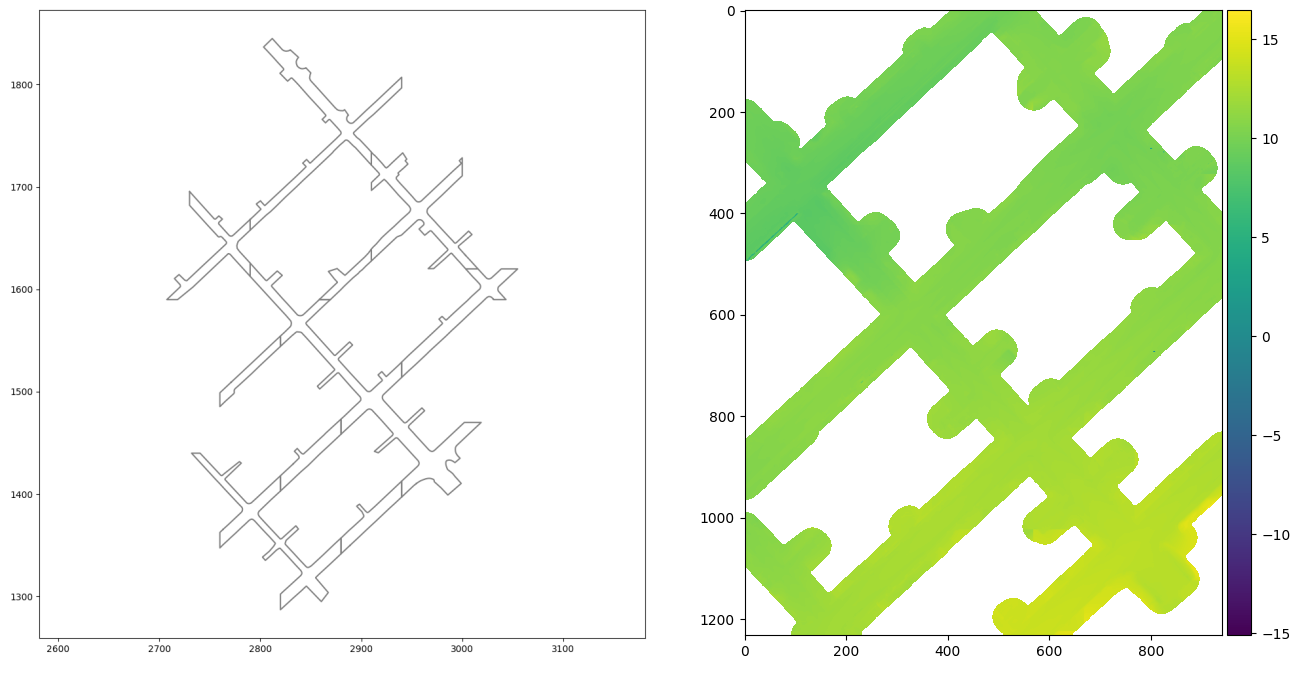}
            \caption{Drivable area vector map (left) and the ground surface height raster map (right)}
		\label{fig:hd-map-b}
	\end{subfigure}
	\caption{The complete HD-map data of a sequence provided in the Argoverse2-TbV dataset.}
	\label{fig:full-hd-map}
\end{figure}

\subsection{Description of HD-map}
\Cref{fig:full-hd-map} shows the complete HD-map as provided in the Argoverse2-TbV dataset. Each sequence contains the HD-map with the following components:
\begin{itemize}
    \item \textbf{Lane graph} (vector map)\\
    The road map constains a lane graph which is the main feature of the HD-map. It provides 3D lane boundaries and additional information, for example, the type of boundary (see \cref{fig:hd-map-a}). 
    \item \textbf{Pedestrian crossings} (vector map)\\
    Each pedestrian crossing is specified by two edgesrepresented by polylines (see \cref{fig:hd-map-a}).
    \item \textbf{Drivable area} (vector map)\\
    The drivable area vector map represents the boundary of known drivable areas as 3D polygon vertices (see \cref{fig:hd-map-b}). 
    \item \textbf{Ground surface height} (raster map)\\
    The ground height is represented as a grid of 30cm resolution in the region of interest. The region of interest is defined within the  5m isocontour of the drivable area (see \cref{fig:hd-map-b}).
\end{itemize}

\subsection{Problem Statement}
Our objective is to obtain a globally consistent map via pose-graph optimization with prior HD-map constraints. Thus, we optimize the following pose-graph cost function:
\begin{align}
    f(\hat{\tf}) &= \sum_{m=0}^{M-1} \norm{\left(\hat{\tf}_{m}^{-1} \hat{\tf}_{m+1}\right) \boxminus\breve{\tf}^{m}_{m+1}}_{\Omega_{m}^1}^2 \nonumber
    \\
    &\qquad\qquad + \sum_{\breve{\tf}_m\in \mathcal{H}} \norm{\hat{\tf}_m^{-1} \boxminus \breve{\tf}_m}_{\Omega_{m}^2}^2, \label{eq: main pgo}
\end{align}
where $\breve{\tf}^{m}_{m+1}$ is the relative pose between two consecutive key frames from the SLAM process (see \Cref{SLAM}), and $\mathcal{H}$ is set of HD-map-registered pose priors (\Cref{PG optimization}).

\section{Methodology}

\begin{figure}[thpb]
	\centering
	\includegraphics[scale=0.51]{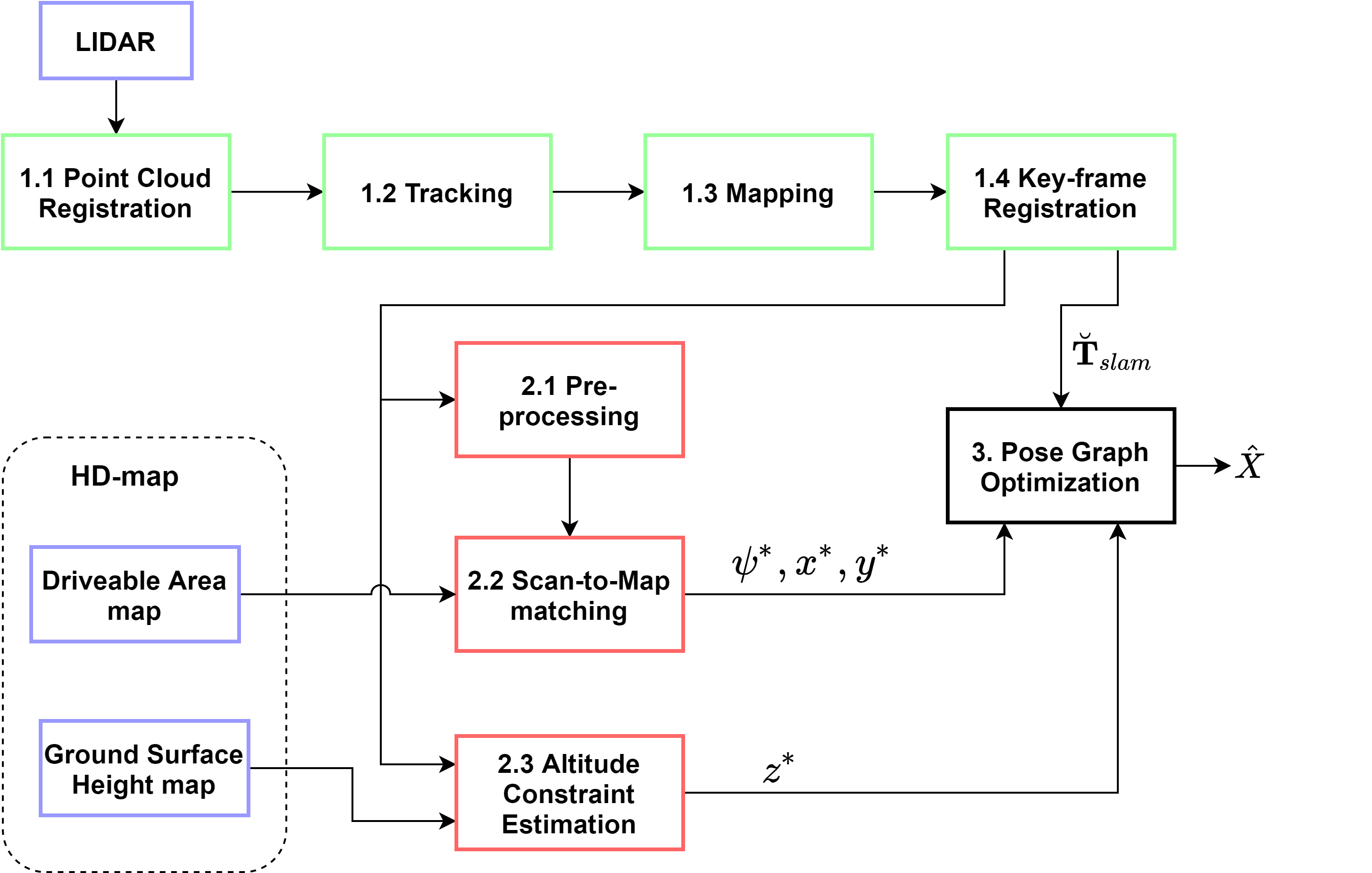}
	\caption{System overview}
	\label{system}
\end{figure}

The complete system overview is provided in the \Cref{system}, where the system is divided into two sections; the SLAM algorithm and the prior map constraint computation. We take the key-frame information from the SLAM algorithm to match with the existing HD-map. Prior map constraints generated from the key-frame to HD-map matching are used for pose-graph optimization.  

\subsection{SLAM} \label{SLAM}
For localization and mapping in our system, we rely on 3D lidar point clouds as sensor data input. We build our algorithm on ALOAM \cite{zhang2017low}, that use corner and planner feature for both tracking and mapping thread. For our system, we optimize their approach by adding a key-frame registration method and pose-graph optimization thread that takes the odometry constraints from the SLAM algorithm and the prior-map constraints. The global map is built after the complete trajectory has been optimized.

\subsection{HD-map to pose prior constraint}

The two HD-map components used in our work to generate prior-map constraints are the ground surface height map and the driveable area vector map (see \Cref{fig:full-hd-map}). We select these two basic elements to make the proposed method generally applicable as they are the basic information available in most HD-maps. 

\subsubsection{Pre-Processing} \label{sec: preprocessing}

The key-frame information includes the point cloud and the pose, used for computing the prior map constraints.

For the pose, which is with respect to the sequence's starting location, we use the transform of this starting location in the prior map to obtain a rough estimate of the key-frame's pose with respect to the prior map, denoted as $\breve{\tf}_\text{slam}$. Given $\breve{\tf}_\text{slam}$, we decompose it into the following:
\begin{equation}
    \breve{\tf}_\text{slam} = \left(\rot(\breve{\psi}) \rot(\breve{\theta}) \rot(\breve{\varphi}), \left[\breve{x},\ \breve{y},\ \breve{z}\right]^\top\right),
\end{equation}
where $\breve{\psi}, \breve{\theta}, \breve{\varphi}$ are the yaw, pitch, roll angles and $\breve{x}, \breve{y}, \breve{z}$ are the translational components. The roll and pitch will be preserved, while the yaw and x, y, z componentes will be refined with the HD map.

To match the point cloud to the HD-map to generate constraints,  we need to extract the \textit{road points} from the key frame point cloud. Since key-frame point cloud is sparse, it is difficult to get enough points that can give a proper road representation from single point clouds. Therefore, we concatenate all frames elapsed between two key-frames to give a more dense point cloud as a key-frame sub-map. We then apply ground segmentation approach proposed in \cite{groundSeg} on the this sub-map points. Once we have the ground plane representation, we employ the road points extraction method from \cite{roadExt}. These road points are then passed on to the prior-map constraint extraction blocks described next.

\subsubsection{Scan-to-Map Matching} \label{sec: scan to map}

For the inclusion of prior information, the challenge is to estimate the prior map constraint from the road points of the key-frame point cloud and the drive-able area vector map.

Regarding the constraint estimation method, for first step, the 2D transformation is computed by scan-to-map matching. The scan is the road points extracted from the key-frame point cloud. The prior map is a set of 2D points generated from the drive-able area vector map. The drive-able vector is given in the form of multiple closed polygons as shown in \Cref{fig:full-hd-map}. Each polygon represents a small area. To use this map for our system, we merge all the polygons given for a sequence to make a single polygon that represent the complete environment. To make a dense representation, we track the polygon boundary at the resolution of 10cm. These points correspond to the curbs on either side of the road. The scan-to-map matching can be done by extracting the curb points from the key-frame point cloud and matching them against the polygon boundary points. Such an approach is susceptible to occlusion from cars parked by the curb or weather conditions. Therefore, we fill the drive-able area polygon with points at 10cm resolution to get a surface representation of the road. This results in more robust and accurate matching using the road points extracted from the key-frame cloud. 

We use Normal Distributions Transform (NDT) \cite{biber2003normal} scan matching to match the key-frame road points with the HD-map points via the following optimization problem on $SE(2)$.
\begin{align}
    &(\psi^*, x^*, y^*)
    = \argmin_{\psi, x, y \in \R} \norm{r_{NDT}(\psi, x, y)}^2_{\Omega_{NDT}}
    \nonumber
    \\
    &\quad\qquad\qquad= \argmin_{\psi, x, y \in \R} \sum_{i=0}^N \norm{r_i(\psi, x, y, (p_i, q_i))}^2_{\Omega_i},
    \\
    &r_i = \begin{bmatrix}\cos(\psi) &-\sin(\psi)\\\sin(\psi) &\cos(\psi)\end{bmatrix}p_i + \begin{bmatrix}x\\y\end{bmatrix} - q_i,\\
    &r = [r_0^\top, \dots r_N^\top]^\top,
\end{align}
where $(p_i, q_i) \in \R^2\times\R^2$ are the set of corresponding point pairs in the prior map and the key-frame road point; $\psi, x, y$ are the yaw and x, y coordinates of the key frame in the prior map. In this work we use the key-frame pose $\breve{\tf}_\text{slam}$ as the initial guess for $\psi, x, y$.
The optimized 2D pose $(\psi^*, x^*, y^*)$ will be used for constructing the 3D pose prior $\breve{\tf}_m \in \mathcal{H}$ introduced in \eqref{eq: main pgo}.
Moreover, we also determine the covariance \cite{bengtsson2003robot, bengtsson2006robust} of the observations $\psi^*, x^*, y^*$  via

\begin{align}
    &\Omega_\text{scan2map} = J^\top \Omega_{NDT} J,\\
    &J \triangleq \left[\frac{\partial r_{NDT}}{\partial \psi},\ \frac{\partial r_{NDT}}{\partial x},\ \frac{\partial r_{NDT}}{\partial y}\right] \bigg\rvert_{\psi^*, x^*, y^*}
\end{align}

\subsubsection{Altitude Constraint Estimation} \label{sec: altitude constraint}

The ground height map is provided in the form of a raster map containing the heights values of the points on the drive-able area in the environment as shown in \Cref{fig:hd-map-b}. The ground surface map gives a general representation of the road height and is a robust way to get the current altitude estimate.
Essentially, the HD-map derived altitude is defined as 
\begin{equation}
    z^* \triangleq g_{SHM}(\breve{x}, \breve{y}),
\end{equation}
where $g_{SHM}(\cdot, \cdot)$ is the ground surface height map function.

\subsection{Pose-Graph Optimization using HD-map Constraints}
\label{PG optimization}

After generating the map constraints, we can assemble the final HD-map pose prior to be used in the pose-graph optimization problem \eqref{eq: main pgo} as follows:

\begin{align}
    &\breve{\tf} = (\breve{\rot}, \breve{\trans}),\\
    &\breve{\rot}=\rot(\psi^*)\rot(\breve{\theta})\rot(\breve{\varphi}),\ \breve{\trans}=\left[x^*, y^*, z^*\right]^\top,
\end{align}
where $\psi^*$, $x^*$, $y^*$ are the scan-to-map matching observations obtained in Sec. \ref{sec: scan to map}, $\breve{\theta}$ and $\breve{\varphi}$ are the roll and pitch estimates from the SLAM module introduced in \ref{sec: preprocessing}, and $z^*$ is the altitude obtained from querying the height map in Sec. \ref{sec: altitude constraint}.

We perform the pose graph optimization using incremental smoothing and mapping (ISAM2) \cite{kaess2012isam2} in GTSAM \cite{dellaert2012factor} library. A priorMapFactor, derived from the NoiseModelFactor1 class, is introduced in addition to the odometry factor to the pose graph in our system. The optimized trajectory is then used for building the map.

\section{Experimental Setup}
In this section we present the baseline methods we compare to, the dataset used for the evaluation and the implementation details of our approach.

\subsection{Methods}

We select three state-of-the-art Lidar based SLAM algorithms to compare the performance of our method against. Our system is build on the LOAM framework. We compare with ALOAM \cite{zhang2017low} and with F-LOAM~\cite{floam}. The latter is an optimized version of LOAM. Furthermore, we also include direct lidar odometry (DLO) \cite{dlo}, which is a direct method as compared to the other feature based approaches. We use the open-source implementations of each method \cite{github_aloam, github_floam, github_dlo} for experiments.

\subsection{Dataset}

We use Argoverse 2: TbV (``Trust, but Verify'') \cite{wilson2023argoverse, lambert2022trust} dataset for the experiments in our work. Argoverse 2 is an autonomous driving dataset with HD-maps, which has been collected across six cities in USA. The TbV dataset contains 1043 sequences with an average duration of 54 seconds, acquired in six different cities. Each sequence consists of sensors data, map data and vehicle poses, and within sensor data lidar point cloud is provided which is collected at 10 HZ from two 32-line lidars. The vehicle poses provided in the dataset are used as ground truth for the evaluation. 

\section{Implementation Details}

We use the LOAM framework as the basis of our SLAM algorithm implementation. For optimization, we add a pose-graph optimization thread using GTSAM \cite{dellaert2012factor} library in our system. As we propose to use constraint computed from the prior map, therefore we implement a custom factor from the NoiseModelFactor1 class provided in GTSAM. We down-sample the dataset by selecting roughly every fifth sequence from the dataset, to reduce the number of sequences while still covering the different locations. This results in a total of 208 sequences. All the experiments are performed on a desktop computer with i7-12700 cpu.

\section{Experiment}

In this section, we  will thoroughly examine the performance of our system, both quantitatively and qualitatively.

\begin{table}
\centering
\caption{Error evaluation for Agrovese2:TBV dataset for ALOAM, F-LOAM, DLO and our method}
\label{tab:agro2_error}
\def\arraystretch{1.15}%
\begin{tabular}{|l|r|r|r|r|}
\hline
       & \multicolumn{1}{l|}{ALOAM \cite{zhang2017low}} & \multicolumn{1}{l|}{F-LOAM \cite{floam}} & \multicolumn{1}{l|}{DLO \cite{dlo}} & \multicolumn{1}{l|}{Our Method} \\ \hline
MEAN   & 0.57                       & 0.80                       & 1.24                     & \textbf{0.25}                   \\ \hline
MEDIAN & 0.45                       & 0.67                       & 1.21                     & \textbf{0.23}                   \\ \hline
RMSE   & 0.75                       & 1.01                       & 1.38                     & \textbf{0.30}                   \\ \hline
StD    & 0.46                       & 0.61                       & 0.54                     & \textbf{0.16}                   \\ \hline
MAX    & 16.36                      & 18.87                      & 28.88                    & \textbf{5.58}                  \\ \hline
\end{tabular}
\end{table}

\begin{figure}[!thpb]
	\centering
	\includegraphics[scale=0.40]{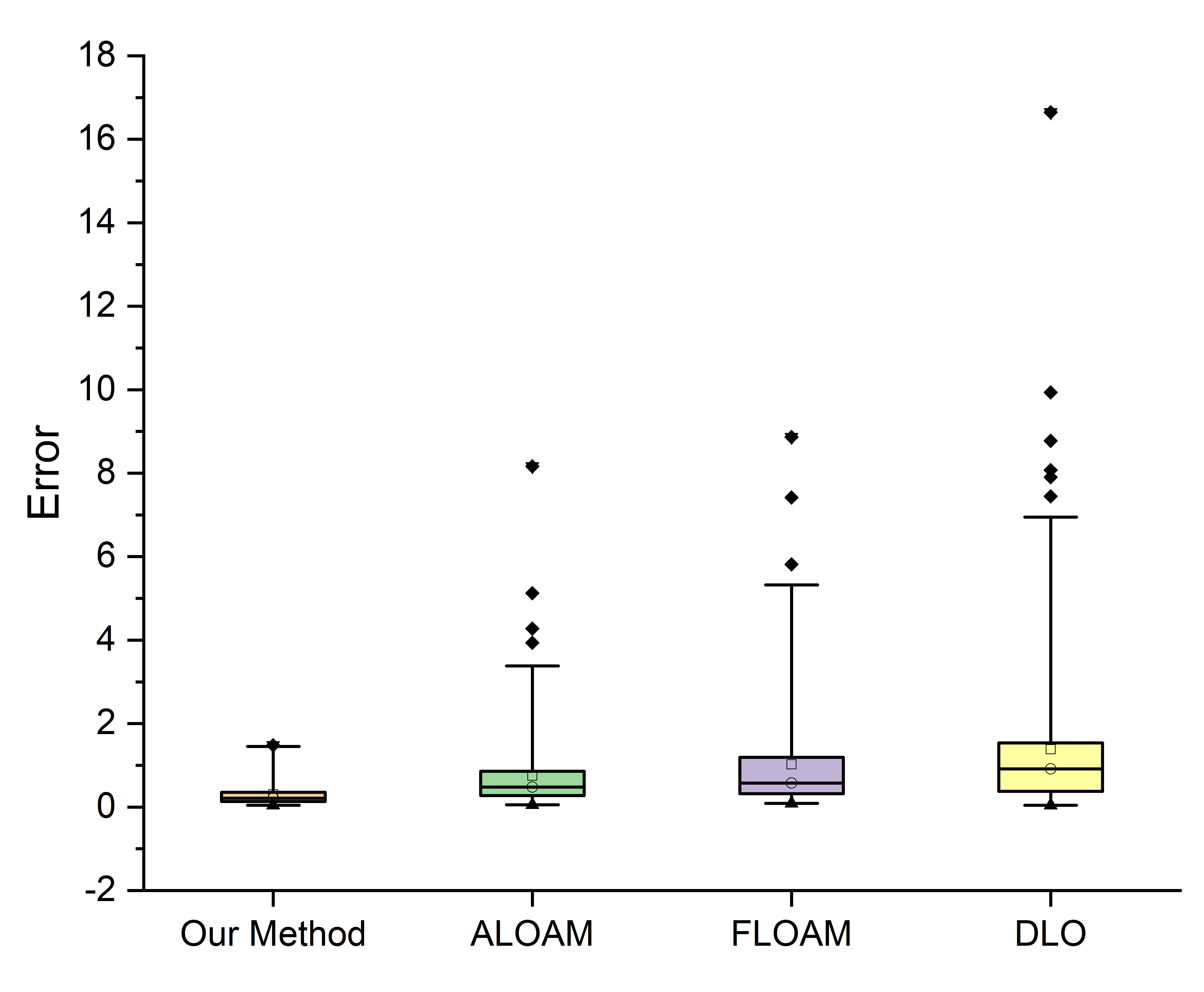}
	\caption{The distribution of RMSE of ATE erros across the 208 sequences from the Agroverse 2 TbV dataset for different algorithms used in our experiments.}
	\label{box_error}
\end{figure}

\subsection{Quantitative Analysis}

In this section, we evaluate the performance of our proposed method against the three baseline SLAM algorithms. We use absolute trajectory error (ATE) \cite{sturm2012benchmark} for evaluation in our work. For an estimated pose $\textbf{P}_i$ and ground truth pose $\textbf{Q}_i$, the absolute pose error $\textbf{E}_i$ can be calculated as follow:

\begin{equation}
    \textbf{E}_i = \textbf{Q}_i^{-1} \textbf{P}_i
\end{equation}

and the RMSE can be estimated as 

\begin{equation}
    \text{RMSE}(\textbf{E}_{1:n}) = \left(\frac{1}{n} \sum_{i=1}^n \|\text{trans}(\textbf{E}_i)\|^2	\right)^{\frac{1}{2}}
\end{equation}

For each sequence, we calculate the abslute trajectory at each frame and then take mean, median, RMSE, standard deviation and the maximum of the errors of the whole sequence. \Cref{tab:agro2_error} shows the average of the error metrics over the 208 sequences. There is a significant improvement over the accuracy from the proposed pose graph optimization method using the prior map constraints. For the RMSE value our method gives an improvement of up to 45cm to the performance of ALOAM, which is a big improvement and make the overall trajectory more consistent. Looking at the results of F-LOAM and DLO, we see much higher error values. These methods have large variations in performance across the dataset. Our proposed approach can produced consistent and accurate results across the 208 sequences.

To get a better understanding of the performance of the four systems across the dataset, we plot the distribution of the RMSE errors for each algorithm as shown in \Cref{box_error}. There is a clear difference in the distributions of the four systems. Starting from right, we see a large deviation in the results of DLO. It can produce high accuracy for some of the sequences but at other scenarios there is significant drop in the performance. Secondly, looking the plot of F-LOAM, it is slightly better than DLO but still with large variations and the maximum error value is more than 8 meters. ALOAM gives better performance in terms of the RMSE distribution. However, its accuracy still deviates in several sequences and the maximum error is similar to F-LOAM. Finally we analyze the performance of our system. It has only one outlier where the value goes above 1m and for majority of the sequences RMSE value is less than 30 cm. 

\begin{figure}[thpb]
	\centering
	\includegraphics[scale=0.40]{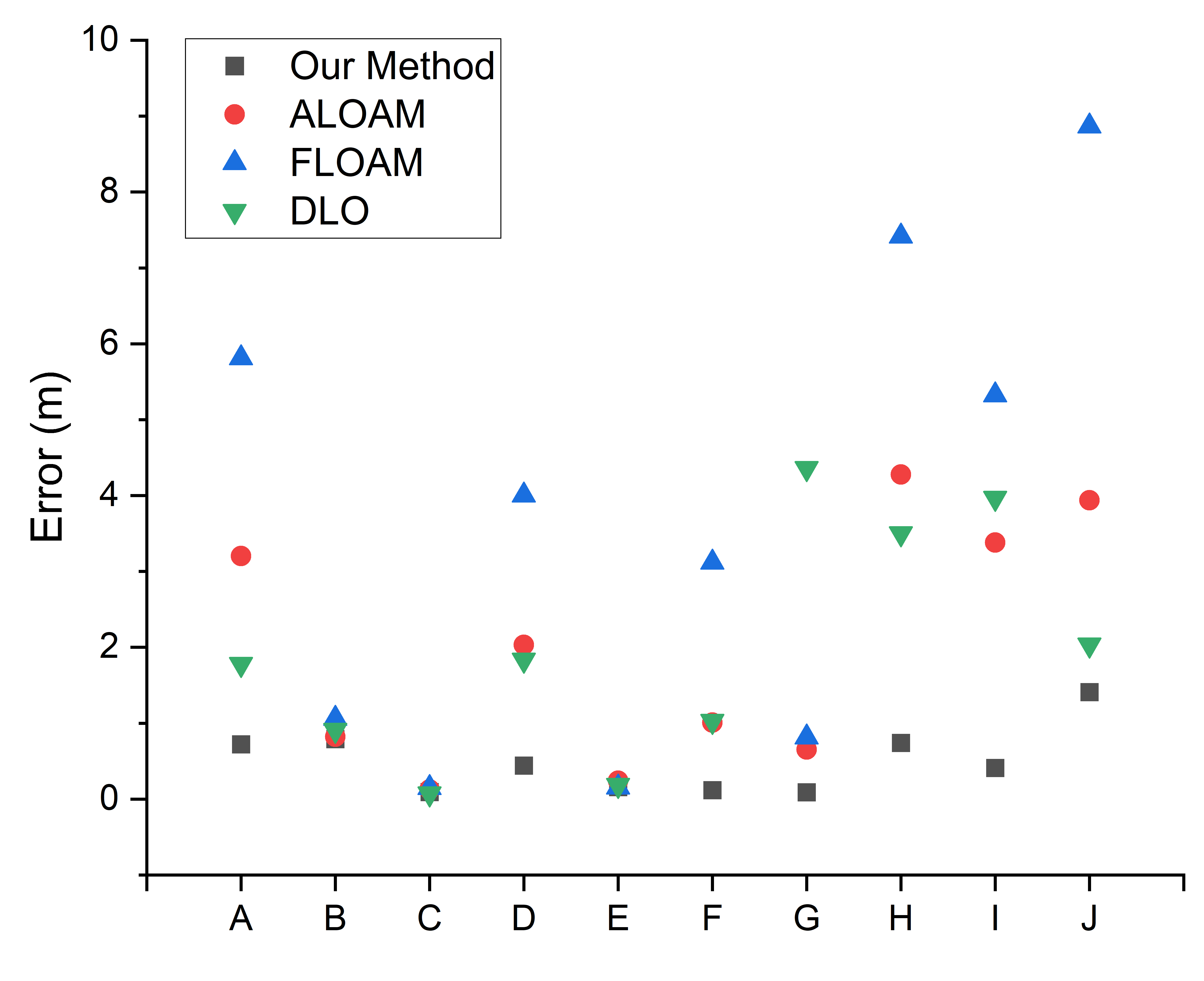}
	\caption{RMSE errors of each method for 10 sequences}
	\label{error_plot}
\end{figure}

\subsection{Qualitative Analysis}
For a more in-depth analysis of the performance of our proposed approach, we examine a few sequences representing different cases from our experiments. \Cref{error_plot} shows the RMSE values of the four methods for 10 sequences. The specific names of these sequences in the dataset are provided in \Cref{tab:seq_names} in the appendix. 

We start by discussing the sequences C and E. For these two cases all the four methods gives very good performance. The difference between the RMSE of our method and the remaining three systems is only a few cm. These are typical results in feature rich environments, where Lidar-based SLAM algorithms give good performance and the prior map optimization can only slightly improve the accuracy. Next, we check on some of the sequences where our approach has resulted in much higher accuracy as compared to baseline SLAM methods. Such cases are the sequences D, F, G, and I. For Sequence D, ALOAM and DLO report errors close 2 m and for F-LOAM it goes up to 4m. But for our method the RMSE value is 0.4 m, which is a significant improvement in accuracy. In the case of sequences F and G, we see that our method reports RMSE of close 10 cm and less than 1 cm respectively, while the remaining methods report much higher errors of more than 1 m. Similarly, we see that our optimization elevate the accuracy in sequence I when Lidar base method have very poor performance. There are some cases, where we saw that when the performance of Lidar based SLAM methods is poor and our system was only able to improve the accuracy to an extent. Two such cases are sequence A and H from the trajectory plots. Both sequences car drives in straight line and both ALOAM and F-LOAM have very high error, while DLO has better performance than the other two method but still has high error. Our system improves the accuracy but final error is around 0.7 m.  

Finally, we want to discuss some limitations of our system, it can be observed from the sequences B and J. For sequence B, our system fails to optimize the trajectory and the resulting RMSE is only a couple of cm less than ALOAM and DLO. Similarly, for sequence J our system gives error of 1.4 m while other systems also give poor results our system was unable to  a more accurate result. The main reason for such limitations is lack of information present in the driveable area vector map for these sequences. The vector map for such scenarios resembles a long corridor resulting in lower accuracy from our system.

\section{CONCLUSIONS}

In this paper, we have presented an approach to use the HD-map components, i.e., driveable area map and ground surface height map as prior information for generating constraint to be used in pose graph optimization. Our approach builds on a typical Lidar base SLAM algorithm and adds key-frame registration strategy. Such key-frames are then used for key-frame scan to HD-map matching to compute the prior map constraints. We performed a thorough evaluation by testing on 208 sequences from Argoverse 2 TbV dataset and compared the performance with state-of-the-art Lidar SLAM methods. Our system proved to give highly accurate results in challenging scenarios where the SOTA lidar based method gave poor performance.




\section*{APPENDIX}
\begin{table}[!htbp]
\caption{Sequence names from the Argoverse2-TbV dataset used in the qualitative analysis.}
\label{tab:seq_names}
\centering
\begin{tabular}{|l|l|}
\hline
  & \multicolumn{1}{c|}{Log Name}                    \\ \hline
A & 6T1aikmJAhnsCKFnyI8Tl1jPn9Kbt73u\_\_Winter\_2021 \\ \hline
B & 7YRWm9Irq4s9v9uwTqF7mly5dIFr4L0O\_\_Autumn\_2020 \\ \hline
C & CcCkXfwbswRCaxcM4d8OkublGV3YlSF4\_\_Autumn\_2020 \\ \hline
D & FfJ9NkcvDVic5SZTP7Irggmg50pX7IRp\_\_Summer\_2020 \\ \hline
E & LEIaZYYGdIH5PxnCG6FcFrc0e5PHGWK5\_\_Autumn\_2020 \\ \hline
F & P4qIURJEN8793diggvoNdfPI3PzfjbCz\_\_Summer\_2020 \\ \hline
G & Qn2tNqunbv80YS0zP1Ns4JvLdHrboSGP\_\_Summer\_2020 \\ \hline
H & cSLEBr2jx2uX7PJWh0RPCUcO6o6Gr1wE\_\_Winter\_2021 \\ \hline
I & fco3TtJKXvY0G6PdHO0MAygGttIGP3e6\_\_Autumn\_2020 \\ \hline
J & nLlUhouz2ynxlMoUdUn7uFR9IAHONnIl\_\_Winter\_2021 \\ \hline
\end{tabular}
\end{table}

\balance
\bibliographystyle{IEEEtran}
\bibliography{references}

\end{document}